\documentclass[letterpaper, 10 pt, conference]{ieeeconf}  

\IEEEoverridecommandlockouts                          
\usepackage{comment}
\usepackage{tabularx}
\usepackage{stfloats}
\usepackage{amsmath, amssymb}
\usepackage{graphicx}
\usepackage{booktabs}
\usepackage{multirow}
\usepackage{caption}
\usepackage{subcaption}
\usepackage{url}
\usepackage{cite}
\usepackage{float}
\usepackage{listings}
\usepackage{xcolor}
\usepackage{hyperref}
\usepackage{pgfplots, tikz}
\pgfplotsset{compat=1.15}
\usepackage{xcolor}
\usepackage{arydshln}
\definecolor{multidiffsensecol}{HTML}{E9CDD4}
\definecolor{pix2pixcol}{HTML}{C7DCEF}
\title{\LARGE \bf MultiDiffSense: Diffusion-Based Multi-Modal Visuo-Tactile Image Generation Conditioned on Object Shape and Contact Pose
}
\author{ Sirine Bhouri*, Lan Wei*, Jian-Qing Zheng, Dandan Zhang
\thanks{*Equal Contribution. Sirine Bhouri, Lan Wei, Dandan Zhang are with the Department of Bioengineering, Imperial-X Initiative, Imperial College London, London, United Kingdom. Jian-Qing Zheng is with CAMS-Oxford Institute, University of Oxford, Oxford, United Kingdom. Corresponding: d.zhang17@imperial.ac.uk}
}

\begin{document}

\maketitle
\begin{abstract}
Acquiring aligned visuo-tactile datasets is slow and costly, requiring specialised hardware and large-scale data collection. Synthetic generation is promising, but prior methods are typically single-modality, limiting cross-modal learning. We present MultiDiffSense, a unified diffusion model that synthesises images for multiple vision-based tactile sensors (ViTac, TacTip, ViTacTip) within a single architecture. Our approach uses dual conditioning on CAD-derived, pose-aligned depth maps and structured prompts that encode sensor type and 4-DoF contact pose, enabling controllable, physically consistent multi-modal synthesis. Evaluating on 8 objects (5 seen, 3 novel) and unseen poses, MultiDiffSense outperforms a Pix2Pix cGAN baseline in SSIM by +36.3\% (ViTac), +134.6\% (ViTacTip), and +64.7\% (TacTip). For downstream 3-DoF pose estimation, mixing 50\% synthetic with 50\% real halves the required real data while maintaining competitive performance ($R^2$: ViTac 0.940 vs. 0.919 real-only; ViTacTip 0.937 vs. 0.982; TacTip 0.784 vs. 0.794). MultiDiffSense alleviates the data-collection bottleneck in tactile sensing and enables scalable, controllable multi-modal dataset generation for robotic applications.
\end{abstract}

\section{Introduction}


Robots require both vision and touch to interact safely and effectively with the physical world, supporting tasks such as object recognition \cite{lin_learning_2019}, texture discrimination \cite{mazid_robotic_2006}, and force estimation \cite{venter_tactile_2017}. Vision provides global, long-range context but is brittle under occlusion and specular reflections, whereas tactile sensing offers local contact geometry, slip, and force cues but is inherently short-range. Combining these modalities enables more robust perception and control in contact-rich tasks.

Among tactile sensing solutions, vision-based tactile sensors (VBTSs) treat touch as an imaging problem: an embedded camera observes a deformable skin under controlled illumination to recover contact geometry and related cues \cite{yuan_gelsight_2017}. This imaging-based mechanism has enabled the development of diverse tactile robotic end-effectors for contact-rich manipulation tasks \cite{he2023tacmms,fan2025magicgripper,zhang2024tacpalm}.
Based on this shared mechanism, VBTS designs can be categorized according to their sensing principles. Following a modality-driven taxonomy \cite{fan2025crystaltac}, we distinguish:
(i) Intensity Mapping Method (IMM), which infers shape or pressure from spatial variations in reflected light \cite{fan2024design};
(ii) Marker Displacement Method (MDM), which measures deformation by tracking printed or embedded markers \cite{lepora_soft_2021};
(iii) Modality Fusion Method (MFM), which employs transparent “see-through” skins and tailored illumination to expose the contact interface and fuse visual appearance with tactile cues \cite{fan2024magictac}.
These sensing principles emphasize complementary physical cues, and many widely used sensors integrate them in different configurations. As a result, spatially and temporally aligned multi-modal datasets are critical for consistent learning and cross-modal generalization across heterogeneous tactile modalities.

In this work, we focus on TacTip (MDM), ViTac (IMM+MFM), and ViTacTip (IMM+MDM+MFM) as representative VBTS modalities for multi-modal data generation. TacTip employs internal markers to measure deformation \cite{lepora_soft_2021}. ViTac removes internal markers and leverages a transparent skin to enable direct visual observation of the contact interface \cite{fan_vitactip_2024}. ViTacTip integrates both mechanisms within a single unit, combining transparent skin and biomimetic markers to synchronize visual and tactile evidence \cite{fan_vitactip_2024}. Related see-through designs further highlight the advantages of exposing the contact interface for multi-modal inference \cite{zhang2025design}. These sensors emphasize complementary cues and therefore suit different tasks: TacTip provides accurate shear and indentation estimates for slip detection; ViTac captures high-fidelity contact appearance and geometry for object and texture recognition; and ViTacTip balances both signals within a unified sensing platform.
Spatial alignment across these modalities enables cross-modality conversion (ViTac$\leftrightarrow$TacTip$\leftrightarrow$ViTacTip), allowing a single generative model to produce the modality required by a downstream task without hardware modification. However, acquiring large-scale aligned datasets across these modalities remains a major bottleneck. Physical tactile data collection is costly, time-consuming, and accelerates sensor wear due to repeated contact cycles \cite{zhong_tactgen_2025,chen_bidirectional_2022}, limiting the scalability of tactile learning and deployment.

To address this bottleneck,  some researchers have pursued synthetic tactile data generation through simulation-based methods that consist in modelling the physics behind sensor-object interaction to simulate a digital version of the sensor and render synthetic tactile images 
\cite{kappassov_simulation_2020, wang_tacto_2022,si_taxim_2022,agarwal_simulation_2021}. However, although these simulators are physically grounded, the generated images often lack realism, exhibiting a significant sim-to-real gap due to the difficulty of accurately modeling soft-body deformations and complex optical effects.
To mitigate this gap, learning-based approaches have emerged that train data-driven generative models to synthesize tactile data. These methods have evolved from conditional GANs \cite{lee_touching_2019, cai_visual-tactile_2021, li_connecting_2019, patel_deep_2020} to conditional diffusion models \cite{higuera_learning_2023, lin_vision-based_2024, luo_controltac_2025}. While these approaches improve visual realism, they remain largely constrained to single sensor modalities.

This single-modality limitation poses a fundamental challenge for tactile sensing research, where robotic platforms often employ diverse sensor configurations tailored to specific applications and hardware constraints. For example, some systems integrate separate visual cameras and tactile sensors, requiring spatially aligned visual–tactile pairs for downstream learning \cite{yang_generating_2023}. Others deploy heterogeneous VBTSs, such as ViTac, TacTip, and ViTacTip, and require aligned data across modalities to enable cross-modal mapping and modality conversion, as demonstrated by Zhang et al. \cite{zhang_design_2025}. However, there is currently no unified generative framework capable of producing spatially aligned and physically consistent synthetic data across heterogeneous VBTSs within a single model. Addressing this gap is essential for scalable multi-modal dataset generation, cross-sensor policy transfer, and flexible deployment across robotic platforms.

To bridge this gap, we present MultiDiffSense, a unified generative framework that synthesizes spatially and temporally aligned ViTac, TacTip, and ViTacTip sensor data within a single architecture.  This work makes three\textbf{ contributions}:




\begin{enumerate}
\item \textbf{Unified generative framework for multi-modal VBTS data.}
We present \emph{MultiDiffSense}, a diffusion-based approach that synthesises \emph{aligned} images for ViTac, TacTip, and ViTacTip within a single model, enabling multi-modal learning and sensor fusion.

\item \textbf{Physically grounded, controllable conditioning.}
Our method conditions on object shape (pose-aligned depth) and contact pose (sensor type and 4-DoF contact), providing geometry-aware control and physically consistent synthesis across heterogeneous sensors.

\item \textbf{Empirical validation across sensors and tasks.}
We evaluate on multiple VBTS families, unseen poses, and novel objects, and demonstrate benefits for downstream pose estimation when synthetic data are mixed with real data.
\end{enumerate}

\section{Related Work}


\subsection{Single-Output Tactile Image Generation}

\subsubsection{Conditional GANs}
Early work framed tactile generation as vision-to-tactile translation with conditional GANs. Lee et al.~\cite{lee_touching_2019} trained bidirectional cGANs on ViTac Cloth, achieving SSIM $\approx 0.9$ but requiring 96,536 aligned samples and focusing on cloth. Li et al.~\cite{li_connecting_2019} scaled to 195 objects yet still needed extensive webcam–GelSight pairing~\cite{yuan_gelsight_2017}. Patel et al.~\cite{patel_deep_2020} used depth image, reaching SSIM $\approx 0.8$ with 578 samples, but validated only on objects with simple features.


\subsubsection{Conditional Diffusion Models}
Diffusion models provide higher fidelity/diversity and more flexible conditioning than GANs. 
Higuera et al.~\cite{higuera_learning_2023} outperformed cGANs on braille classification (75.74\% vs.\ 31.18\%); however, because their model lacked physical conditioning (e.g., force or contact masks), it benefited from additional fine-tuning on real data.
Lin et al.~\cite{lin_vision-based_2024} incorporated force signals, and Luo et al.~\cite{luo_controltac_2025} proposed ControlTac, which leverages ControlNet~\cite{zhang_adding_2023} to generate tactile images from force data, contact masks, and a reference tactile image. These physical priors improve controllability and realism for state-of-the-art single-modality generation. 
However, all approaches remain single-modality, preventing the generation of aligned multi-modal datasets needed for robust fusion-based perception.


\subsection{Multi-Modal Tactile Image Generation}
Multi-modal sensing, especially combining vision and touch, often outperforms single modalities in robotics. Such systems require datasets in which modalities are \emph{spatially and temporally aligned} to capture the same interaction. However, existing aligned resources (e.g., ObjectFolder~2.0~\cite{gao_objectfolder_2022}, ViTac~\cite{luo_vitac_2018}, Touch-and-Go~\cite{yang_touch_2022}) remain limited in scale and coverage relative to contemporary vision corpora, constraining robust fusion and cross-sensor generalisation. This limitation motivates the development of multi-modal data generation methods capable of synthesizing aligned visuo–tactile observations conditioned on object geometry and contact pose. Such methods can augment training data, facilitate cross-modality conversion, and reduce reliance on costly real-world data collection.




Extending generative models from single to multi-modal synthesis to produce large aligned datasets poses key challenges: (i) temporal alignment across sensors with different rates and noise; (ii) cross-modal physical consistency (e.g., visual slip should correlate with tactile shear); and (iii) a unified conditioning representation, since features salient in one modality may not transfer to another.
Training separate models per modality scales as $\mathcal{O}(N)$ and cannot guarantee cross-modal physics. 
Sequential conditioning approaches~\cite{zhong_touching_2023}, where one modality is generated first and others conditioned on it, mitigate some issues but suffer from error propagation and neglect the inherently bidirectional nature of multi-sensory relationships




\section{Methods}
We address the task of multi-sensor modality image generation for robotic perception, where 
the goal is to synthesise TacTip, ViTac, and ViTacTip sensor outputs under precise geometric and spatial control.

\subsection{Preliminary}

Latent Diffusion Models (LDMs) \cite{rombach_high-resolution_2022} are a type of diffusion models that operate in the latent space of a pre-trained autoencoder $D(E(.))$ where $E$ is the encoder and $D$ is the decoder. Stable Diffusion (SD) is an LDM conditioned on text. It is composed of a Vector Quantised-Variational AutoEncoder (VQ-VAE), a time-conditioned U-Net denoising network, and a CLIP text encoder that maps a text prompt into a textual embedding condition $C_\text{text}$ \cite{wang_disco_2024}.
During training, given an image $I$ and text condition $C_\text{text}$, the encoded image latent $z_0 = E(I)$ undergoes diffusion over T timesteps where noise sampled from a pure Gaussian distribution $\epsilon \backsim N(0,1)$ is gradually applied to it to produce the noisy latent $z_T$. The SD model learns the reverse denoising process via the following training objective:
\vspace{-0.1cm}
\begin{equation}
L = E_{z_0,t, C_\text{text},\epsilon} ||\epsilon - \epsilon_\theta(z_t, t, C_\text{text})||^2_2
\end{equation}
where $t = {1,..,T}$ is the diffusion timestep and $\epsilon_\theta$ is the predicted noise.
SD has a U-Net architecture which
accepts the noisy latent $z_t$ and the text embedding condition $C_\text{text}$, as input. After training, a deterministic sampling process (e.g., DDIM \cite{song_denoising_2022})can be applied, to generate $z_0$, the denoised latent and pass it through the decoder $D$ to generate the final image.

ControlNet \cite{zhang_adding_2023} extends SD to allow conditioning the diffusion process on a control image beyond just text prompt. To achieve this, it creates a trainable copy of SD's encoder and middle blocks that can process the control image (e.g. depth maps, edge maps etc). The output of each block is then fed into the original UNET through zero-convolution layers. and inject this geometric information into the generation process. By including an additional condition $C_\text{image}$, the diffusion model's learning objective therefore becomes:
\begin{equation}
L = E_{z_0,t, C_\text{text},C_\text{image},\epsilon} ||\epsilon - \epsilon_\theta(z_t, t, C_\text{text},C_\text{image} )||^2_2
\end{equation}

\begin{figure*}[!th]
\centering \includegraphics[width=0.9\linewidth]{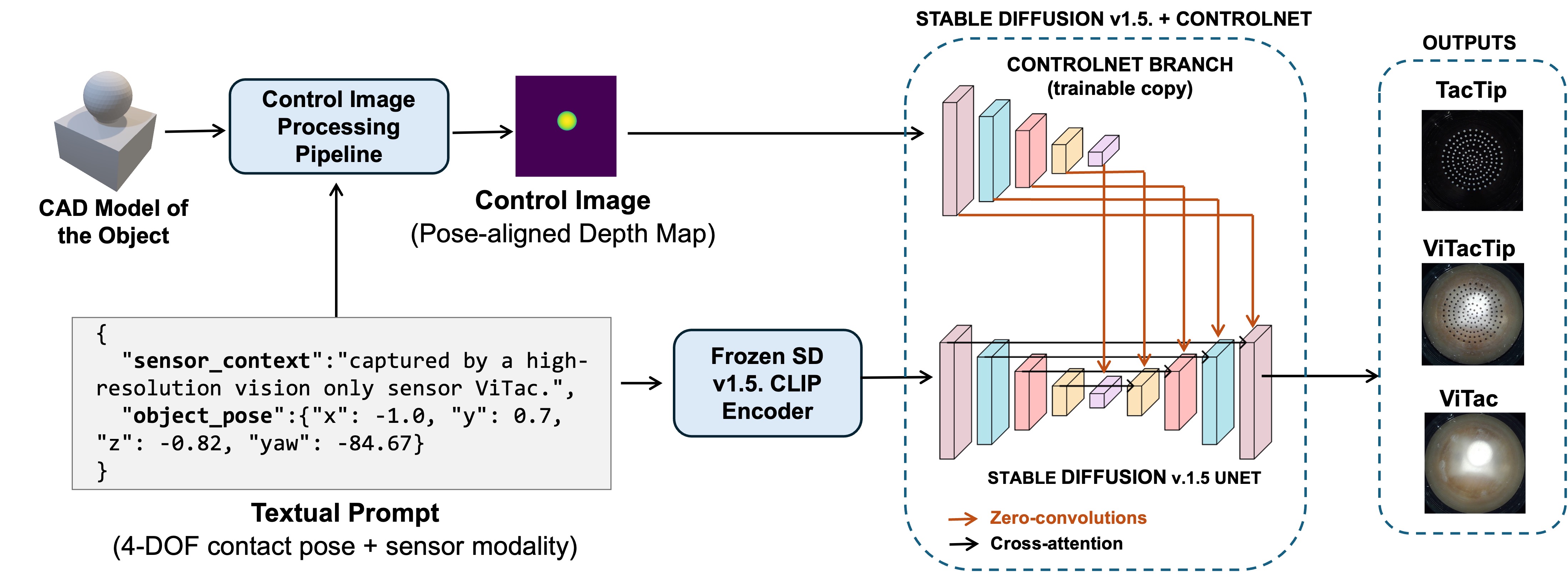}
\vspace{-0.2cm}
\captionsetup{font=footnotesize,labelsep=period}
\caption{Framework Overview.
The model takes a CAD file and textual prompt as inputs. The CAD model is converted into a pose-aligned depth map (control image) fed via zero-convolutions into the ControlNet branch as the geometric condition. The text prompt is encoded with CLIP and injected into the UNet via cross-attention. The decoder then refines the latents based on both conditions to generate an image reflecting the desired object geometry, contact pose, and sensor modality.}
\label{fig:overview_detailed}
\vspace{-0.3cm}
\end{figure*}

\subsection{Model Architecture}
MultiDiffSense builds on the ControlNet framework integrated with SD v1.5 to allow dual conditioning on textual prompts and geometric depth maps. An overview of MultiDiffSense's framework is shown in Fig.~\ref{fig:overview_detailed}.

\textbf{Model Input:} Our method takes two inputs: (1) a structured textual prompt $C_\text{text}$ that specifies the sensor modality $m \in \{\text{TacTip}, \text{ViTac}, \text{ViTacTip}\}$ and contact pose $p$ defined by 4 degrees of freedom $(x,y,z, \theta_\text{z})$, and (2) a control image $C_\text{image} \in \mathbb{R}^{H \times W}$ containing a pose-aligned depth map rendered from the CAD model at pose $p$, where $H$ and $W$ are the height and width of the image, respectively.
The contact pose parameters are defined in the sensor-centred coordinate frame with the $z$-axis pointing outward from the sensor surface as: $x,y \in [-5,5]$ mm representing horizontal displacement from the sensor centre, $z \in [-1,1]$ mm representing indentation depth, and $\theta_z \in [-90°, 90°]$ representing yaw rotation about the sensor's $z$-axis. Our objective is to learn a generator $G_\theta$ that models the conditional distribution $P(I_m \mid C_\text{text}, C_\text{image})$, where $I_m \in \mathbb{R}^{H \times W \times 3}$ is the generated RGB tactile sensor image for modality $m$ given the conditions $C_\text{text}$ and $C_\text{image}$.

This dual conditioning allows the model to be guided by both semantic properties (via text
prompts) and geometric configuration (via CAD-derived depth maps) with textual conditioning ($C_\text{text}$) mainly functioning as a modality-selection mechanism that supports unified multi-sensor generation, while depth map conditioning ($C_\text{image}$) ensures realism and spatial alignment. Importantly, because the
4-DoF pose $p$  in $C_\text{text}$ corresponds exactly to the object pose in $C_\text{image}$, the model learns a cross-modal mapping between language and spatial layout, enabling accurate and controllable image synthesis without requiring force readings, contact masks, or reference tactile images.  
Images are first encoded into a latent space ($64 \times 64 \times 4$) via a variational autoencoder (VAE). The U-Net denoising network operates within this latent space, gradually refining noisy latents over multiple timesteps before decoding back to full $512 \times 512$ pixel images. Multi-scale attention layers facilitate interactions between text, geometry, and latent features.

\textbf{Textual pathway:} Structured prompts $C_\text{text}$ are encoded using a pre-trained CLIP text encoder, producing a 512-dimensional embedding. These embeddings are injected via cross-attention at multiple U-Net levels, providing fine-grained semantic and modality-specific guidance.  

\textbf{Geometric pathway:} The raw CAD model of the desired object goes through a processing pipeline to generate a depth map that is aligned with the 4-DoF pose $p$ given in the textual prompt. The resulting CAD-derived depth map $C_\text{image}$ is then fed into a parallel ControlNet encoder 
branch and the obtained feature maps are injected into the main 
SD v1.5 UNET via zero-convolutions. This ensures that harmful noise is not added to the 
deep features of the pre-trained SD v1.5 model at the beginning of training and 
therefore protects the trainable copy from being damaged. The depth maps provide structural
constraints independent of sensor artefacts, enabling the model to gradually learn
geometry-consistent image generation.  


\begin{figure*}
    \centering
    \includegraphics[width=0.85\linewidth]{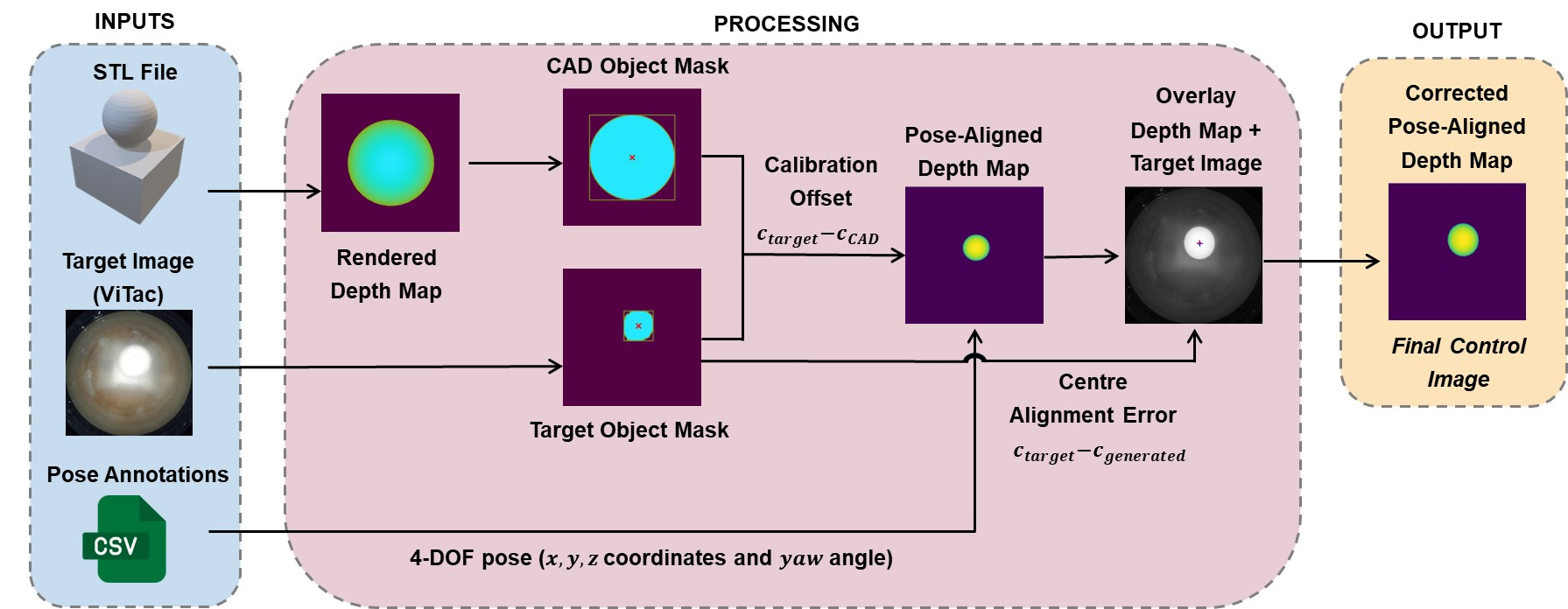}
    \captionsetup{font=footnotesize,labelsep=period}
    \caption{Control Image Processing Pipeline.
The pipeline takes an STL file, target image and a CSV log of end-effector poses (pose annotations) as inputs and consists of four stages: 
(1) Use STL file to render depth map and
preprocess it to extract clean object masks; 
(2) Align robot coordinates to image pixels via centroid
mapping; 
(3) Scale XY translations using workspace calibration, Incorporate Z-axis depth through
geometric scaling and intensity modulation, and Apply yaw rotation using 2D rotation matrices; 
(4) Centre alignment error is minimised to $<$ 5 pixels ($\approx$0.6\,mm)}
\label{fig:control_processing}
\vspace{-0.3cm}
\end{figure*}

\textbf{Conditions fusion.} At inference time, to combine unconditional and dual-conditioned 
predictions, classifier-free guidance is employed as per the original implementation of ControlNet ~\cite{zhang_adding_2023}, allowing 
control over adherence to conditioning while maintaining generative diversity:  
\vspace{-0.1cm}
\begin{equation}
\epsilon_{\text{pred}} = \epsilon_{\text{uncond}} + w_{\text{cfg}} 
\left( \epsilon_{\text{cond}} - \epsilon_{\text{uncond}} \right),
\end{equation}
\vspace{-0.1cm}
where $\epsilon_{\text{pred}}$ is the final model output, $\epsilon_{\text{uncond}}$ is the 
unconditional noise prediction, $\epsilon_{\text{cond}}$ is the conditional prediction 
incorporating both text and control conditioning, and $w_{\text{cfg}}$ is the guidance 
weight controlling conditioning strength. 

\subsection{Data Conditioning Pipeline}
\subsubsection{Control Image Generation}
To generate control images $C_\text{image}$ for ControlNet training, we developed a multi-stage pipeline that transforms CAD models into pose-aligned depth maps with geometric consistency validation. The pipeline addresses coordinate system ambiguity, implements adaptive calibration, and incorporates error correction feedback. A detailed diagram can be found on Fig. \ref{fig:control_processing}.

\subsubsection{Textual Prompt Generation}
Structured textual prompts $C_\text{text}$ were written in JSON format to capture both semantic and spatial information. They included both the 4-DoF contact pose and the desired sensor modality. An example prompt is shown in Fig.~\ref{fig:json_prompt_short}.

\begin{figure}[!t]
\centering
\begin{minipage}{\linewidth}
\lstset{
  basicstyle=\ttfamily\footnotesize,
  breaklines=true,
  breakatwhitespace=true
}
\begin{lstlisting}
{
  "sensor_context": "captured by a high-resolution vision-based tactile sensor ViTac.",
  "object_pose": {"x": 3.17, "y": 0.97, "z": -0.49, "yaw": 89.9}
}
\end{lstlisting}
\end{minipage}
\vspace{-0.25cm}
\captionsetup{font=footnotesize,labelsep=period}
\caption{Example of Structured Textual Prompt}
\label{fig:json_prompt_short}
    \vspace{-0.5cm}
\end{figure}

\section{Experiments and Results}
\subsection{Dataset Introduction}
We train and test the model on the ViTacTip, TacTip and ViTac datasets \cite{fan_vitactip_2024}. The collection process consisted in mounting each sensor as the end effector of the Dobot MG400 desktop arm and collecting data as the contact poses were varied from $[-5, -5, -1, -90]$ to $[5, 5, 1, 90]$ with $[X (\text{mm}), Y (\text{mm}), Z (\text{mm}), \theta (^\circ)]$. For each object–sensor pair, 500 images were collected. 
 
To build our dataset, five objects with different geometric complexity and contact patterns were selected from the original datasets \cite{fan_vitactip_2024}: straight edge (linear), cuboid (planar), sphere (curved), Pacman shape (mixed convex/concave), and hollow cylinder (internal/external curvature). This yielded 2,500 samples per modality and 7,500 total
(i.e., 500 frames × 5 objects × 3 modalities). Poses $p$ were synchronised across sensors to ensure aligned multi-modal samples.
For each object, we generate pairs of pose-aligned depth maps and structured text prompts for each modality matched to corresponding ground-truth tactile images during training and testing.


\subsection{Experimental Setup}
We adopt a stratified 70/15/15 train-validation-test split to ensure robust evaluation while preserving cross-modal correspondence. In total, 5,250 samples are used for training, 1,125 for validation, and 1,125 for testing (corresponding to 1,750/375/375 per modality). Splits are performed at the (object, pose) level such that, for any given object-pose pair, the corresponding TacTip, ViTac, and ViTacTip images are assigned to the same partition. This strategy preserves spatial alignment, enables learning of cross-modal relationships, and prevents data leakage across splits.
 
\textbf{MultiDiffSense Training.} All experiments were implemented in PyTorch 1.10/Python 3.9 and trained on a single NVIDIA A100 (80\,GB, CUDA 12.0) with $512{\times}512$ inputs. 
We used AdamW ($\text{lr}{=}1{\times}10^{-5}$), DDIM with a linear noise schedule, and batch size 8. 
Early stopping (patience{=}10) governed training (max 78{,}840 steps). 
Following ControlNet~\cite{zhang_adding_2023}, we initialise from SD v1.5: the original U-Net is frozen; a parallel ControlNet branch is initialised with the same pre-trained weights; and the zero-convolution layers linking ControlNet to the U-Net are zero-initialised to stabilise training while preserving pre-trained generative capacity.

 
\textbf{Baseline Model Training.} 
We adopt Pix2Pix cGANs~\cite{isola_image--image_2018} as the baseline, following Fan et al.~\cite{fan_vitactip_2024}. 
Models are trained on identical splits with the same depth-map conditioning as MultiDiffSense. 
Because cGANs lack text-prompt conditioning, we train three separate models (TacTip, ViTac, ViTacTip), each mapping depth to its target modality. 
Training uses vanilla adversarial loss plus $L_1$ reconstruction ($\lambda{=}100$), batch size 8, and $256{\times}256$ inputs. 
Each model is trained for 300 epochs with an initial learning rate of $2{\times}10^{-4}$ for 200 epochs, linearly decayed to 0 over the final 100 epochs.

\subsection{Evaluation Metric}
We assess generation quality with five complementary metrics. Pixel fidelity is measured by MSE and PSNR between generated and ground-truth images. 
Structural fidelity uses SSIM, capturing local luminance, contrast, and structure relevant to contact geometry. 
Perceptual similarity is evaluated with LPIPS, and distributional realism with FID computed on feature distributions of real vs.\ generated sets.
For downstream utility, we assess pose prediction accuracy on held-out real tactile data using MSE, RMSE, MAE, and $R^2$ over $(X, Z, \theta_z)$, measuring how well synthetic images preserve the geometric information required for robotic perception.

\subsection{Main Results}
\subsubsection{Seen Objects (Unseen Poses)}
We evaluate our MultiDiffSense framework on its ability to generalise to unseen contact poses for objects encountered during training. As shown in Table \ref{tab:seen_unseen_pose}, MultiDiffSense demonstrates strong performance across all three sensor modalities, significantly outperforming the Pix2Pix cGAN baseline. Our method achieves higher SSIM (0.919, 0.877, 0.768 for ViTac, ViTacTip, TacTip) and substantially lower LPIPS and FID, confirming the superior perceptual and distributional quality of our generated images.
 
However, performance varies across sensor modalities, and this variation appears to correlate with the level of abstraction required to model each modality. ViTac, which primarily captures visual cues related to object appearance, shape, and pose, achieves the highest SSIM scores (0.919 for seen objects and 0.912 for unseen objects). This is expected given its more direct geometric correspondence with the input depth maps.
In contrast, TacTip yields lower SSIM scores (0.768 and 0.741, respectively), reflecting the increased difficulty of synthesizing purely tactile deformation patterns, which exhibit a more indirect relationship to geometric depth information.
ViTacTip demonstrates intermediate performance (0.877 and 0.835), balancing the geometric clarity of visual cues with the additional structural complexity introduced by tactile markers.

\begin{table}[!t]
\centering
\captionsetup{font=footnotesize,labelsep=period}
\caption{Seen objects, unseen poses: performance across tactile modalities (ViTac, ViTacTip, TacTip). We compare MultiDiffSense (single unified model) with Pix2Pix cGAN (separate per modality). Metrics are mean±std; ↑/↓ denote higher-/lower-better; best per metric–modality in bold.}
\renewcommand{\arraystretch}{1.1}
\resizebox{\linewidth}{!}{
\begin{tabular}{c:c:ccc}
\hline\hline
\textbf{Metric} & \textbf{Model} & \textbf{ViTac} & \textbf{ViTacTip} & \textbf{TacTip} \\
\hline
\multirow{2}{*}{SSIM $\uparrow$} 
 & MultiDiffSense   & \textbf{0.919 $\pm$ 0.022} & \textbf{0.877 $\pm$ 0.024} & \textbf{0.768 $\pm$ 0.058} \\
 & Pix2Pix cGAN & 0.678 $\pm$ 0.028 & 0.362 $\pm$ 0.015 & 0.450 $\pm$ 0.012 \\
\hline
\multirow{2}{*}{PSNR $\uparrow$} 
 & MultiDiffSense   & \textbf{28.27 $\pm$ 3.10}  & \textbf{25.74 $\pm$ 2.06}  & \textbf{22.61 $\pm$ 2.33} \\
 & Pix2Pix cGAN & 20.57 $\pm$ 0.92  & 17.38 $\pm$ 0.242 & 14.87 $\pm$ 0.28 \\
\hline
\multirow{2}{*}{MSE $\downarrow$} 
 & MultiDiffSense   & \textbf{0.002 $\pm$ 0.003} & \textbf{0.003 $\pm$ 0.002} & 0.006 $\pm$ 0.004 \\
 & Pix2Pix cGAN & 0.009 $\pm$ 0.003 & 0.018 $\pm$ 0.001  & 0.033 $\pm$ 0.002 \\
\hline
\multirow{2}{*}{LPIPS $\downarrow$} 
 & MultiDiffSense   & \textbf{0.091 $\pm$ 0.031} & \textbf{0.059 $\pm$ 0.012} & \textbf{0.141 $\pm$ 0.035} \\
 & Pix2Pix cGAN & 0.285 $\pm$ 0.028 & 0.251 $\pm$ 0.010 & 0.235 $\pm$ 0.016 \\
\hline
\multirow{2}{*}{FID $\downarrow$} 
 & MultiDiffSense  & \textbf{17.287} & \textbf{2.212} & \textbf{26.021 }\\
 & Pix2Pix cGAN & 175.505 & 46.417 & 93.445 \\
\hline\hline
\end{tabular}}
\label{tab:seen_unseen_pose}
\end{table}

\subsubsection{Unseen Objects}
To evaluate the ability of MultiDiffSense to generalise to completely novel objects, a critical requirement for real-world deployment, we tested models on three objects unseen during training (300 samples total, 100 per object). As shown in Table \ref{unseen_results}, MultiDiffSense maintains robust performance. While metrics show expected degradation compared to seen objects (e.g., SSIM dropping from 0.877 to 0.835 for ViTacTip), the model generalises relatively well across all modalities. The performance hierarchy across sensors remains consistent, with ViTac achieving the best results (SSIM: 0.912) and TacTip the most challenging (SSIM: 0.741).
 
\begin{figure*}[!th]
\centering
\captionsetup{font=footnotesize,labelsep=period}
\includegraphics[width=1\linewidth]{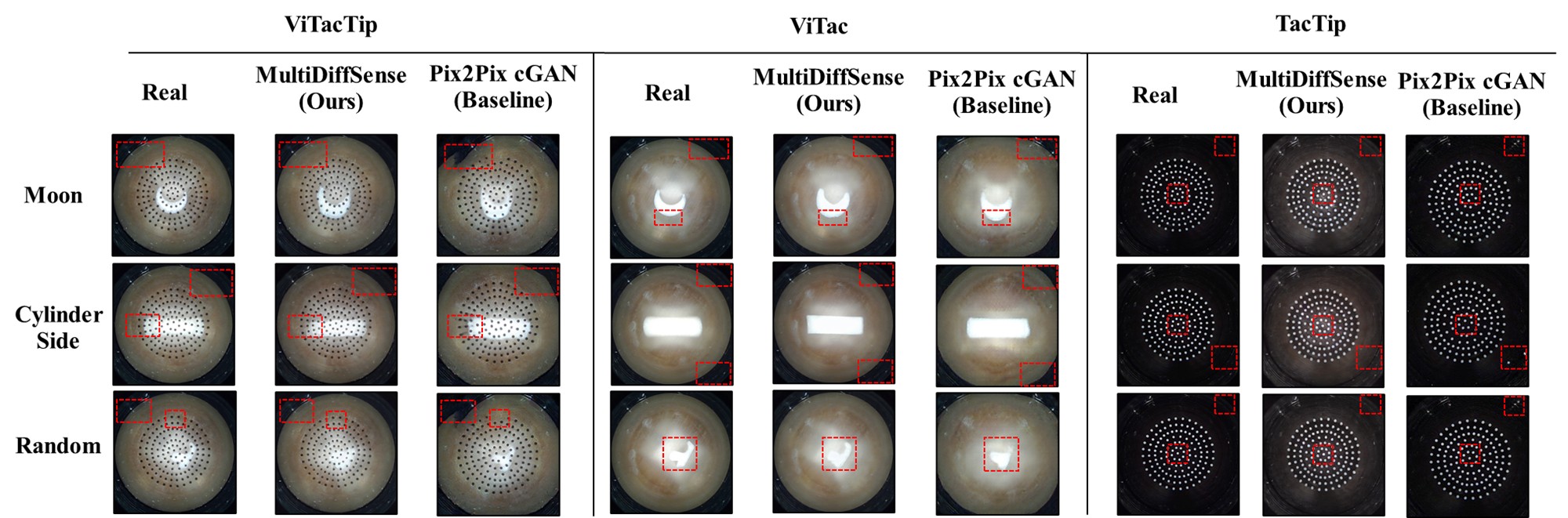}
\caption{Visualisation of image generation result on unseen objects across three tactile sensor modalities (ViTacTip, ViTac, TacTip). 
Red dashed boxes highlight regions where the methods differ: MultiDiffSense better preserves contact geometry, marker patterns, and lighting.}
\label{fig:unseen_object_figure}
\vspace{-0.1cm}
\end{figure*}

Critically, our framework substantially outperforms the Pix2Pix cGAN baseline across all metrics and modalities. Fig. \ref{fig:unseen_object_figure} and Table \ref{unseen_results} illustrate this performance gap clearly, with MultiDiffSense achieving an averaged SSIM of 0.829 across the three modalities compared to the baseline's 0.492, and a lower averaged LPIPS (0.114 vs 0.268).
\begin{table}[t]
\centering
\captionsetup{font=footnotesize,labelsep=period}
\caption{Quantitative results on \emph{unseen objects} across three tactile modalities (ViTac, ViTacTip, TacTip). Best per metric–modality in bold.}
\renewcommand{\arraystretch}{1.1} 
\resizebox{\linewidth}{!}{
\begin{tabular}{c:c:ccc}
\hline\hline
\textbf{Metric} & \textbf{Model} & \textbf{ViTac} & \textbf{ViTacTip} & \textbf{TacTip} \\
\hline
\multirow{2}{*}{SSIM $\uparrow$} 
 & MultiDiffSense   & \textbf{0.912 $\pm$ 0.013} & \textbf{0.835 $\pm$ 0.030} & \textbf{0.741 $\pm$ 0.066} \\
 & Pix2Pix cGAN & 0.669 $\pm$ 0.015 & 0.356 $\pm$ 0.012 & 0.450 $\pm$ 0.012 \\
\hline
\multirow{2}{*}{PSNR $\uparrow$} 
 & MultiDiffSense   & \textbf{27.314 $\pm$ 2.146}  & \textbf{23.530 $\pm$ 1.608}  & \textbf{21.296 $\pm$ 2.575} \\
 & Pix2Pix cGAN & 20.165 $\pm$ 0.462  & 17.163 $\pm$ 0.217  & 14.871 $\pm$ 0.271 \\
\hline
\multirow{2}{*}{MSE $\downarrow$} 
 & MultiDiffSense   & \textbf{0.002 $\pm$ 0.001} & \textbf{0.005 $\pm$ 0.002} & \textbf{0.009 $\pm$ 0.006} \\
 & Pix2Pix cGAN & 0.009 $\pm$ 0.001 & 0.019 $\pm$ 0.001 & 0.033 $\pm$ 0.002 \\
\hline
\multirow{2}{*}{LPIPS $\downarrow$} 
 & MultiDiffSense   & \textbf{0.116 $\pm$ 0.019} & \textbf{0.074 $\pm$ 0.015} & \textbf{0.152 $\pm$ 0.040} \\
 & Pix2Pix cGAN & 0.307 $\pm$ 0.020 & 0.255 $\pm$ 0.010 & 0.241 $\pm$ 0.018 \\
\hline
\multirow{2}{*}{FID $\downarrow$} 
 & MultiDiffSense   & \textbf{60.342} & \textbf{2.798} & \textbf{28.833} \\
 & Pix2Pix cGAN & 185.995 & 55.459 & 90.946 \\
\hline\hline
\end{tabular}}

\label{unseen_results}
\vspace{-0.3cm}
\end{table}

\begin{table*}[h]
\centering
\captionsetup{font=footnotesize,labelsep=period}
\caption{Pose estimation performance comparison across sensor modalities and training dataset types.
Mixed datasets achieve performance comparable to or superior to real-only training.
Best results for each row (per component) for each metric are shown in bold.}
\resizebox{\textwidth}{!}{%
\begin{tabular}{c|c|c|c|c|c|c|c|c|c|c|c|c|c}
\hline\hline
\multirow{2}{*}{\textbf{Modality}} & \multirow{2}{*}{\textbf{Component}} & \multicolumn{4}{c|}{\textbf{Real Dataset}} & \multicolumn{4}{c|}{\textbf{Mixed Dataset}} & \multicolumn{4}{c}{\textbf{Synthetic Dataset}} \\
\cline{3-14}
& & \textbf{MSE $\downarrow$} & \textbf{RMSE $\downarrow$} & \textbf{MAE $\downarrow$} & \textbf{R² $\uparrow$} & \textbf{MSE $\downarrow$} & \textbf{RMSE $\downarrow$} & \textbf{MAE $\downarrow$} & \textbf{R² $\uparrow$} & \textbf{MSE $\downarrow$} & \textbf{RMSE $\downarrow$} & \textbf{MAE $\downarrow$} & \textbf{R² $\uparrow$} \\
\hline
\multirow{3}{*}{\textbf{TacTip}} 
& $X$ & \textbf{3.607} & \textbf{1.899} & \textbf{1.500} & \textbf{0.610} & 4.333 & 2.081 & 1.636 & 0.532 & 6.182 & 2.486 & 1.992 & 0.332 \\
& $Z$ & 0.066 & 0.258 & 0.221 & 0.789 & \textbf{0.028} & \textbf{0.166} & \textbf{0.129} & \textbf{0.912} & 0.279 & 0.528 & 0.475 & 0.112 \\
& $\theta_{z}$ & \textbf{42.527} & \textbf{6.521} & \textbf{5.532} & \textbf{0.982} & 221.682 & 14.889 & 5.700 & 0.907 & 602.863 & 24.553 & 13.818 & 0.748 \\
\hline
\multirow{3}{*}{\textbf{ViTac}} 
& $X$ & 0.183 & 0.428 & 0.309 & 0.980 & \textbf{0.131} & \textbf{0.361} & \textbf{0.280} & \textbf{0.986} & 0.903 & 0.950 & 0.761 & 0.902 \\
& $Z$ & 0.068 & 0.261 & 0.213 & 0.782 & \textbf{0.051} & \textbf{0.226} & \textbf{0.183} & \textbf{0.837} & 0.594 & 0.770 & 0.689 & -0.893 \\
& $\theta_{z}$ & 15.130 & 3.890 & 3.037 & 0.994 & \textbf{6.755} & \textbf{2.599} & \textbf{2.123} & \textbf{0.997} & 15.041 & 3.878 & 2.900 & 0.994 \\
\hline
\multirow{3}{*}{\textbf{ViTacTip}} 
& $X$ & \textbf{0.171} & \textbf{0.413} & \textbf{0.314} & \textbf{0.982} & 0.348 & 0.590 & 0.436 & 0.962 & 14.716 & 3.836 & 3.031 & -0.591 \\
& $Z$ & \textbf{0.010} & \textbf{0.102} & \textbf{0.085} & \textbf{0.967} & 0.047 & 0.217 & 0.181 & 0.850 & 0.143 & 0.378 & 0.299 & 0.545 \\
& $\theta_{z}$ & 4.432 & 2.105 & 1.703 & \textbf{0.998} & \textbf{4.250} & \textbf{2.062} & \textbf{1.751} & \textbf{0.998} & 37.678 & 6.138 & 4.699 & 0.984 \\
\hline\hline
\end{tabular}
}

\label{tab:pose_estimation}
\vspace{-0.4cm}
\end{table*}

\subsubsection{Comparative Analysis}
\textbf{Advantages over Pix2Pix Baseline:}
Our MultiDiffSense demonstrates superior performance through two key architectural advantages. First, generation quality; visual inspection reveals that cGAN-generated images suffer from substantial blur and noise artefacts, particularly affecting object boundaries. In contrast, MultiDiffSense produces sharper, more realistic tactile patterns that better preserve geometric information crucial for downstream robotic tasks. This most likely stems from the iterative denoising process that allows gradual refinement through multiple steps, compared to cGANs' single-step generation that struggles to bridge the substantial semantic gap between geometric depth maps and complex sensor images.
 
Second, background consistency: cGANs exhibit severe deformation of sensor background regions (Fig. \ref{fig:unseen_object_figure}) as the generator prioritises foreground object generation, leading to inconsistent spatial reconstruction. MultiDiffSense benefits from SD's extensive pre-training on natural images, providing rich structural priors that maintain spatial coherence (i.e. undeformed background) and promote smoother image structures through the denoising objective.
 
\textbf{Advantages over Existing Cross-Modal Approaches:} 
Compared to existing cross-modal tactile sensing methods using Pix2Pix cGANs like Fan et al. \cite{fan_vitactip_2024}, MultiDiffSense offers significant practical advantages through its unified architecture. Where traditional approaches require training separate Pix2Pix cGANs for each cross-modal conversion task, our single conditional diffusion model handles all three modalities through text-based specification. This unified approach provides two key benefits: (1) reduced training complexity and computational overhead by eliminating multiple model training, and (2) inherent scalability since incorporating new sensor modalities requires only adjusting textual conditioning rather than training entirely new conversion models for each sensor pair.
These advantages collectively explain why MultiDiffSense achieves superior performance across all evaluation scenarios, demonstrating its potential for practical robotic applications requiring high-fidelity multi-modal tactile image generation.
\subsubsection{Pose Estimation Downstream Task}
To assess the realism and utility of generated images, we evaluated synthetic data on pose estimation, a representative robotic task that tests whether synthetic tactile data retains fine-grained geometric information necessary for downstream robotic tasks such as tactile servoing \cite{lepora_pose-based_2021}.
Fan et al \cite{fan_vitactip_2024} showed that images collected using ViTac, ViTacTip and TacTip sensors can be used to train a ResNet18 model to achieve accurate edge pose regression. Following their evaluation protocol, we conducted pose regression experiments where models estimate the sensor's pose relative to a cylindrical edge from tactile images. If tactile images generated by MultiDiffSense preserve sufficient geometric and contact information, they should enable successful pose estimation comparable to real data \cite{fan_vitactip_2024, kendall_posenet_2016}.
 
We trained ResNet18 to estimate three pose parameters: horizontal displacement $X$ from the sensor centre, indentation depth $Z$, and yaw angle $\theta_{z}$ about the $Z$-axis. Our dataset comprised 500 tactile images per sensor modality with pose values sampled within [-5, 5] mm for $X$, [-1, 1] mm for $Z$, and [-90, 90] degrees for $\theta_{z}$. We used 80\% for training and 20\% for testing.
Training employed photometric data augmentation (grayscale conversion, sharpness adjustment, colour jitter, and Gaussian blur) while avoiding geometric transformations that would alter the ground truth pose labels. Models were trained for 100 epochs using AdamW optimiser ($\text{lr}{=}1{\times}10^{-4})$, with $L_1$ loss and a batch size of 8.
 
Three training regimes were compared: 100\% real dataset, 100\% synthetic dataset, and a mixed dataset with 50\% real data and 50\% synthetic data. For each regime, three separate models were trained, one for each sensor modality, to enable separate evaluation of the diffusion model's generation quality for each sensor type.
 
The results shown in Table \ref{tab:pose_estimation} demonstrate the feasibility of using MultiDiffSense for tactile data augmentation while revealing important sensor-specific performance variations. Mixed datasets frequently achieve performance comparable to or superior to real-only training, particularly evident in ViTac's X-displacement (0.361mm vs 0.428mm) and TacTip's Z-displacement estimation (0.166mm vs 0.258mm). This suggests that adding synthetic data to the training dataset introduces cleaner representations of the underlying geometric relationships between tactile inputs and object poses, preventing the model from overfitting to sensor-specific noise in real data.
 
However, purely synthetic training shows degraded performance, with TacTip's yaw estimation being most severely affected (24.553° vs 6.521° for real data). This indicates that while synthetic images contain sufficient geometric information for effective data augmentation, complete replacement of real tactile data remains challenging, particularly for VBTS with strictly tactile sensing where complex deformation patterns are difficult to synthesise accurately.


\subsection{Ablation Studies}
\subsubsection{Effect of the additional geometric condition}
To evaluate the impact of dual conditioning versus single conditioning, we trained two model variants: (1) control-only using geometric conditioning alone through the control image, and (2) dual conditioning combining textual prompts with control image. Both variants were trained using identical architecture and hyperparameters, with test results averaged over three independent runs and reported in Table \ref{tab:ablation_control}.
To isolate the contribution of each conditioning configuration while minimising computational overhead, the ablation variants were trained exclusively on a single modality (ViTacTip), unlike our final model which leverages all three sensor modalities.
 
The results from Table \ref{tab:ablation_control} reveal comparable performance between control-only and dual-conditioned variants. On seen objects, the dual-conditioned model shows marginal improvement (control→dual: $\Delta$SSIM +0.001, $\Delta$FID -0.443). However, on unseen objects, the control-only variant demonstrates slight superiority (control→dual: $\Delta$SSIM +0.008, $\Delta$FID -0.642). Given the limited number of runs and observed variances, these differences are insufficient to establish systematic superiority of either approach.
The marginally lower performance of the dual-conditioned (2) model on unseen objects likely stems from the increased complexity of reconciling two conditioning inputs with novel data, representing a more challenging generalisation task.
 
These findings confirm that geometric conditioning (control image) serves as the dominant factor in tactile sensor image generation, while semantic conditioning (textual prompts) provides supplementary but meaningful contributions. This aligns with our task's inherently geometric nature, where object shape and pose are paramount. Importantly, prompt conditioning becomes essential for multi-modal generation, as it provides the mechanism to distinguish between sensor modalities and enables targeted generation of specific sensor types at inference time.

\begin{table}[!t]
\centering
\captionsetup{font=footnotesize,labelsep=period}
\caption{Ablation 1: Impact of geometric Control (CAD-derived depth) on MultiDiffSense. Evaluated on \emph{seen objects–unseen poses} and \emph{unseen objects}.
Best results are bolded. Metrics are mean$\pm$std over three runs.}
\resizebox{\linewidth}{!}{
\begin{tabular}{c:cc:cc}
\hline\hline
\multirow{2}{*}{\textbf{Metric}} & \multicolumn{2}{c:}{\textbf{Seen Objects}} & \multicolumn{2}{c}{\textbf{Unseen Objects}} \\
\cmidrule(lr){2-3} \cmidrule(lr){4-5}
 & Control-only & Prompt + Control & Control-only & Prompt + Control \\
\hline
SSIM $\uparrow$ & 0.852 $\pm$ 0.001 & \textbf{0.853 $\pm$ 0.001} & \textbf{0.820 $\pm$ 0.001} & 0.812 $\pm$ 0.001 \\
FID $\downarrow$ & 2.464 $\pm$ 0.050 & \textbf{2.021 $\pm$ 0.035} & \textbf{3.224 $\pm$ 0.274} & 3.866 $\pm$ 0.021 \\
\hline\hline
\end{tabular}
}

\label{tab:ablation_control}
\end{table}

 \begin{table}[!t]
\centering
\captionsetup{font=footnotesize,labelsep=period}
\caption{Ablation 2: Impact of prompt length on reconstruction quality for \emph{seen} and \emph{unseen objects}. Best results per case are bolded.}
\resizebox{\linewidth}{!}{
\begin{tabular}{c:cc:cc}
\hline\hline
\multirow{2}{*}{\textbf{Metric}} & \multicolumn{2}{c:}{\textbf{Seen Objects}} & \multicolumn{2}{c}{\textbf{Unseen Objects}} \\
\cmidrule(lr){2-3} \cmidrule(lr){4-5}
 & Short Prompt (1) & Long Prompt (2) & Short Prompt (1) & Long Prompt (2) \\
\hline
SSIM $\uparrow$   & \textbf{0.8768} & 0.8394 & \textbf{0.8349} & 0.8069 \\
PSNR $\uparrow$   & \textbf{25.74} & 23.91 & \textbf{23.53} & 22.24 \\
MSE $\downarrow$  & \textbf{3.04E-03} & 4.53E-03 & \textbf{4.81E-03} & 6.29E-03 \\
LPIPS $\downarrow$ & \textbf{0.0593} & 0.0765 & \textbf{0.0738} & 0.0893 \\
FID $\downarrow$  & 2.212 & \textbf{2.1993} & \textbf{2.7982} & 3.5497 \\
\hline\hline
\end{tabular}
}
\label{tab:prompt_length}
\vspace{-0.2cm}
\end{table}

\begin{figure}[!t]
    \centering
    \captionsetup{font=footnotesize,labelsep=period}
    \includegraphics[width=1\linewidth]
    {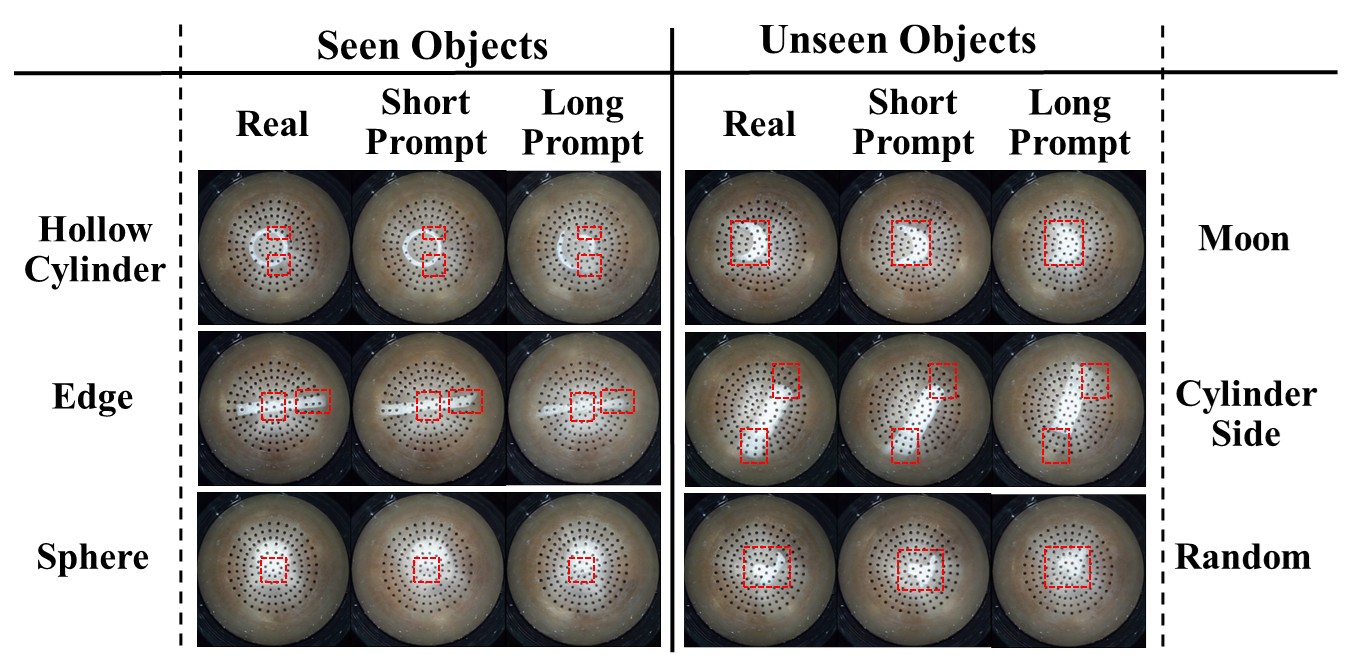}
    \caption{Effect of prompt length on reconstruction quality. 
    Real vs Generated images by the two different model variants under the two testing scenarios (seen
    object but unseen poses and unseen objects).}
\label{fig:Ablation_results2}
    \vspace{-0.3cm}
\end{figure}

\subsubsection{Effect of the structure of the textual prompt}
We investigated how prompt complexity affects generation quality, comparing minimal short prompts (Fig. \ref{fig:json_prompt_short}) to longer comprehensive prompts which include more fields: ``object\_description'', ``contact\_description'', ``sensor\_context'', ``style\_tags'', ``negatives'' and ``object\_pose''. 
 
As shown on Table \ref{tab:prompt_length} and Fig. \ref{fig:Ablation_results2}, short prompts consistently outperform long prompts across all metrics and test scenarios. For seen objects, short prompts achieve superior SSIM (0.877 vs 0.839), lower LPIPS (0.059 vs 0.077), and better PSNR (25.74 vs 23.91 dB), with similar advantages maintained for unseen objects.
Indeed, short prompts reduce the parameter space the model must learn to map, creating a more constrained optimisation problem that is easier to solve with limited training data (5,250 samples across three modalities). On the other hand, the comprehensive descriptions in long prompts may introduce conflicting or redundant information that complicates the learning process.
 
However, with larger, more diverse datasets containing varied objects, materials, and contact scenarios, long prompts would theoretically provide better conditioning signal for generating more finely-controlled tactile images. The current results suggest that prompt complexity should be matched to dataset scale and diversity; minimal prompts for constrained datasets, comprehensive prompts for rich, large-scale data that can support complex semantic conditioning.

\section{Conclusions and Future Work}
MultiDiffSense is the first unified framework to generate spatially and temporally aligned tactile data across multiple sensor modalities within a single diffusion model. Conditioning on geometric control images and structured textual prompts enables controllable synthesis across ViTac, TacTip, and ViTacTip while preserving alignment for cross-modal learning.
Our method outperforms a single-modality cGAN baseline (SSIM: +36.3\%, +134.6\%, +64.7\% on unseen objects for ViTac, ViTacTip, TacTip) while consolidating three models into one.


Future work will focus on scaling the framework to larger and more geometrically diverse object sets to further enhance generalisation. Extending the approach to complex object categories, including articulated and deformable objects, represents an important step toward broader real-world applicability. Incorporating richer geometric and material representations beyond depth maps may further improve synthesis fidelity for transparent, reflective, or texture-dominant surfaces.
Another promising direction is expanding the current 4-DoF contact parameterisation to full 6-DoF interaction modelling and temporal sequence generation, enabling the synthesis of dynamic contact events such as slip, rolling, and continuous manipulation. Such extensions would support learning policies for contact-rich manipulation under realistic temporal dynamics.

\bibliographystyle{IEEEtran}
\bibliography{references}

\end{document}